\documentclass[letterpaper]{article} 
\usepackage{aaai2026}  
\usepackage{times}  
\usepackage{helvet}  
\usepackage{courier}  
\usepackage[hyphens]{url}  
\usepackage{graphicx} 
\usepackage{natbib}  
\usepackage{caption} 
\usepackage[ruled,vlined,linesnumbered,noend]{algorithm2e}
\usepackage{booktabs}
\usepackage{siunitx}
\usepackage[table]{xcolor}
\usepackage{verbatim}

\usepackage{amsmath, amssymb}

\usepackage{newfloat}
\usepackage{listings}
\DeclareCaptionStyle{ruled}{labelfont=normalfont,labelsep=colon,strut=off} 
\title{Multimodal Robust Prompt Distillation for 3D Point Cloud Models}
\author{
    Xiang Gu\textsuperscript{\rm 1}\thanks{These authors contributed equally.}, 
    Liming Lu\textsuperscript{\rm 1}\footnotemark[1], 
    Xu Zheng\textsuperscript{\rm 2,\rm 3}\thanks{Corresponding author.}, 
    Anan Du\textsuperscript{\rm 4}, 
    Yongbin Zhou\textsuperscript{\rm 1}, 
    Shuchao Pang\textsuperscript{\rm 1}\footnotemark[2]
    \\
}
\affiliations{
    \textsuperscript{\rm 1}Nanjing University of Science and Technology\\ 
    \textsuperscript{\rm 2}The Hong Kong University of Science and Technology (Guangzhou)\\
    \textsuperscript{\rm 3}INSAIT, Sofia University, St. Kliment Ohridski\\ 
    \textsuperscript{\rm 4}Nanjing University of Industry Technology\\
    \{eminentguxiang, luliming, zhouyongbin, pangshuchao\}@njust.edu.cn, zhengxu128@gmail.com, anan.du@niit.edu.cn
}

\begin{document}

\maketitle
\begin{abstract}
Adversarial attacks pose a significant threat to learning-based 3D point cloud models, critically undermining their reliability in security-sensitive applications. Existing defense methods often suffer from (1) high computational overhead and (2) poor generalization ability across diverse attack types.
To bridge these gaps, we propose a novel yet efficient teacher-student framework, namely \textbf{M}ultimodal \textbf{R}obust \textbf{P}rompt \textbf{D}istillation (MRPD) for distilling robust 3D point cloud model.
It learns lightweight prompts by aligning student point cloud model's features  with robust embeddings from three distinct teachers: a vision model processing depth projections, a high-performance 3D model, and a text encoder.
To ensure a reliable knowledge transfer, this distillation is 
guided by a confidence-gated mechanism which dynamically balances the contribution of all input modalities. Notably, since the distillation is all during the training stage, there is no additional computational cost at inference. Extensive experiments demonstrate that MRPD substantially outperforms state-of-the-art defense methods against a wide range of white-box and black-box attacks, while even achieving better performance on clean data.
Our work presents a new, practical paradigm for building robust 3D vision systems by efficiently harnessing multimodal knowledge.
\end{abstract}
\begin{links}
     \link{Code}{https://github.com/eminentgu/MRPD}
\end{links}

\section{Introduction}

\begin{figure}[ht!]
    \centering
    \includegraphics[width=0.9\linewidth]{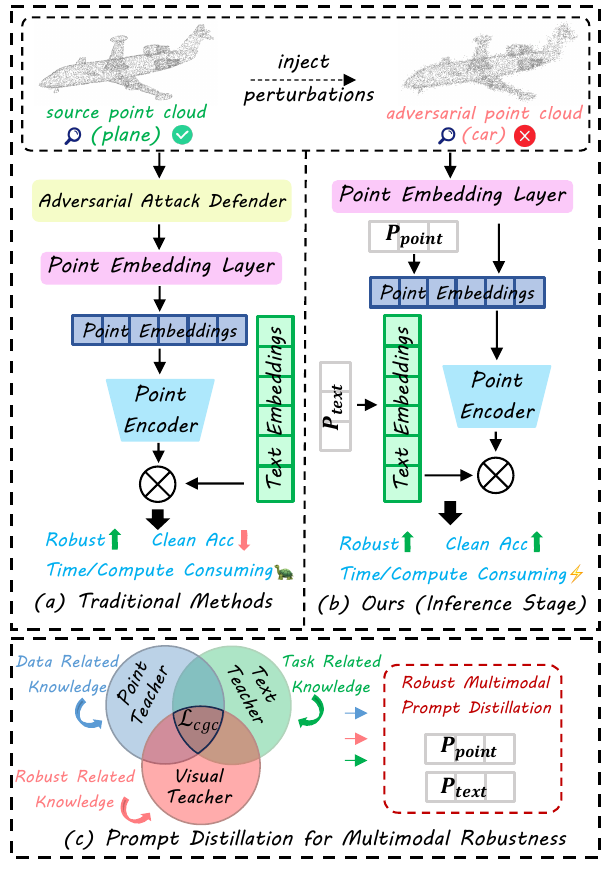}
    \caption{
    (a) Traditional Defenses: Heavy modules with increasing inference costs. (b) Our Inference: The optimized prompts provide robustness with \textbf{zero} computational overhead. (c) Our Training: We distill robust, multimodal knowledge into lightweight prompts.}
    \vspace{-8pt}
    \label{fig1}
\end{figure}

Deep learning\cite{lu2024uniads} has significantly advanced 3D perception, enabling autonomous systems, such as self-driving cars, to understand their environments with impressive accuracy~\cite{sohail2024advancing}.
However, these models remain highly vulnerable to adversarial attacks~\cite{zhang20233d, pang2025towards}, where small, often imperceptible perturbations cause severe prediction failures. This fragility undermines their reliability in open-world settings and raises serious safety concerns in high-stakes applications. For example, in autonomous driving, a maliciously altered point cloud could lead to misidentifying a pedestrian, with potentially fatal consequences~\cite{wang2021adversarial}. 

This vulnerability is especially pronounced in 3D point clouds. Although they provide rich geometric cues essential for robotics and autonomous vehicles, their sparse and unstructured nature creates a broad attack surface, making them particularly susceptible to adversarial manipulation~\cite{xiao2018generating, liu2019extending}.
In response, various defense methods have been proposed~\cite{wicker2019robustness, wu2020if, zhang2023art, zhang2024maffn}, but they still face a fundamental trade-off. Many are highly specialized and fail to generalize across attacks, while more robust methods, such as adversarial training or architectural modifications, sincur significant computational costs. This makes them unsuitable for real-time, latency-sensitive scenarios\cite{wang2024dp,pang2025pridm,lu2025ciard}. As illustrated in Figure~\ref{fig1}(a), these conventional defenses often rely on cumbersome modules that increase inference latency, limiting their practical deployment.

To break this paradigm of relying on computationally expensive, single-modality defenses, recent advances in large-scale Vision Language Models (VLMs)\cite{xia2025one} offer a compelling alternative. 
These models have demonstrated remarkable robustness and knowledge transfer capabilities, inspiring efforts to adapt their pretrained knowledge to the 3D domain, with methods like PointCLIP\cite{zhang2022pointclip, zhu2023pointclip} and ULIP\cite{xue2023ulip} showing great promise. 
However, these pioneering efforts have overwhelmingly focused on boosting standard task accuracy, largely overlooking the crucial opportunity to transfer the robustness of these powerful models.
Consequently, a systematic mechanism to endow 3D perception with the robustness advantages of 2D vision, without incurring the heavy costs of traditional defenders, remains largely absent.

To address these challenges, we propose a novel yet efficient teacher-student framework: Multimodal Robust Prompt Distillation (MRPD). 
Instead of relying on costly adversarial examples or altering the model's architecture, our MRPD distills robustness into lightweight, learnable prompts and further improve robustness of 3D point cloud models. 
As shown in Figure~\ref{fig1}(c), during the training phase, we leverage a combination of three robust ``teachers'': (1) a frozen image encoder that provides stable supervision, as 3D adversarial perturbations often lose efficacy when projected to 2D; (2) a text encoder guided by learnable prompts to discover a more robust semantic space; and (3) a powerful point cloud teacher model that offers high-quality geometric guidance on clean data. 
This multimodal knowledge from these distinct teachers is transferred to the student model's learnable prompts through a carefully designed distillation process, featuring a confidence-gated mechanism to filter unreliable teacher signals and a dynamic weighting strategy to balance the different knowledge sources.

Crucially, the complex distillation process is discarded after training. As shown in Figure~\ref{fig1}(b), only the lightweight, optimized prompts are retained, providing adversarial resilience with \textbf{zero} additional computational overhead.
Our main contributions are summarized as follows:
(I) We propose Multimodal Robust Prompt Distillation (MRPD), a novel framework that efficiently transfers multimodal robustness from image, text, and teacher models into lightweight prompts, thereby leaving the student model architecture untouched.
(II) We propose a highly efficient paradigm that achieves robust defense with zero inference overhead, effectively breaking the longstanding trade-off between adversarial robustness and clean data accuracy.
(III) Through extensive experiments, we demonstrate that MRPD establishes a new state-of-the-art in 3D adversarial defense, consistently outperforming computationally expensive methods against a diverse suite of attacks.

\section{Related Work}

\noindent\textbf{Adversarial Defense for Point Clouds.}
In response to this diverse threat landscape, a similarly varied array of defense mechanisms has been proposed for 3D point cloud models. 
A common strategy involves pre-processing the input point cloud to remove or repair adversarial perturbations. 
This includes early methods like Statistical Outlier Removal (SOR\cite{rusu2008towards}) and their more advanced successors, such as DUP-Net\cite{zhou2019dup} and IF-Defense\cite{wu2020if}, which focus on input reconstruction and purification. Rather than pre-processing inputs at inference time, another line of thought focuses on augmenting the training data to build more intrinsically robust models. 
For instance, PointGuard\cite{liu2021pointguard} utilizes random subsampling with a majority voting scheme, while Point-CutMix\cite{zhang2022pointcutmix} employs a mixup-based strategy for regularization. 
More recently, diffusion models like Ada3Diff\cite{zhang2023ada3diff} have emerged as powerful tools for purifying adversarially perturbed inputs. 
Despite their effectiveness against certain threats, these defense paradigms often suffer from two major drawbacks: limited generalizability across attack types and significant computational overhead. By contrast, our MRPD charts a different course by aiming for a general-purpose, efficient defense that enhances the model's intrinsic robustness without costly pre-processing or architectural changes at inference time.

\noindent\textbf{Vision-Language Guided Point Cloud Models.}
Beyond these traditional defense paradigms, a promising new direction emerges from leveraging the rich knowledge within large-scale VLMs like CLIP\cite{radford2021learning}. 
Initial approaches, such as PointCLIP\cite{zhang2022pointclip} and PartSLIP\cite{liu2023partslip}, sought to bridge the 2D and 3D domains by projecting point clouds into multi-view depth maps, thereby capitalizing on powerful pre-trained 2D encoders. 
To avoid the geometric information loss inherent in 2D projection, a second type of models emerged. 
These models, including CLIP2Point\cite{huang2023clip2point} and ULIP\cite{xue2023ulip}, focus on directly aligning 3D encoders with the VLM feature space using large-scale text-image-point cloud triplet datasets, a paradigm that recent works like UNI3D\cite{zhou2023uni3d} have pushed to billion-parameter scales. 
However, a critical blind spot in this line of research has been its overwhelming focus on improving performance on standard downstream tasks (\textit{e.g.}, classification), leaving the crucial opportunity to transfer robustness from the 2D domain largely unexplored. 
Our work directly addresses this gap, proposing a systematic teacher-student framework to distill this untapped multimodal knowledge into the 3D models.

\section{Methodology}
\begin{figure*}[t]
    \centering
    \includegraphics[width=0.98\linewidth]{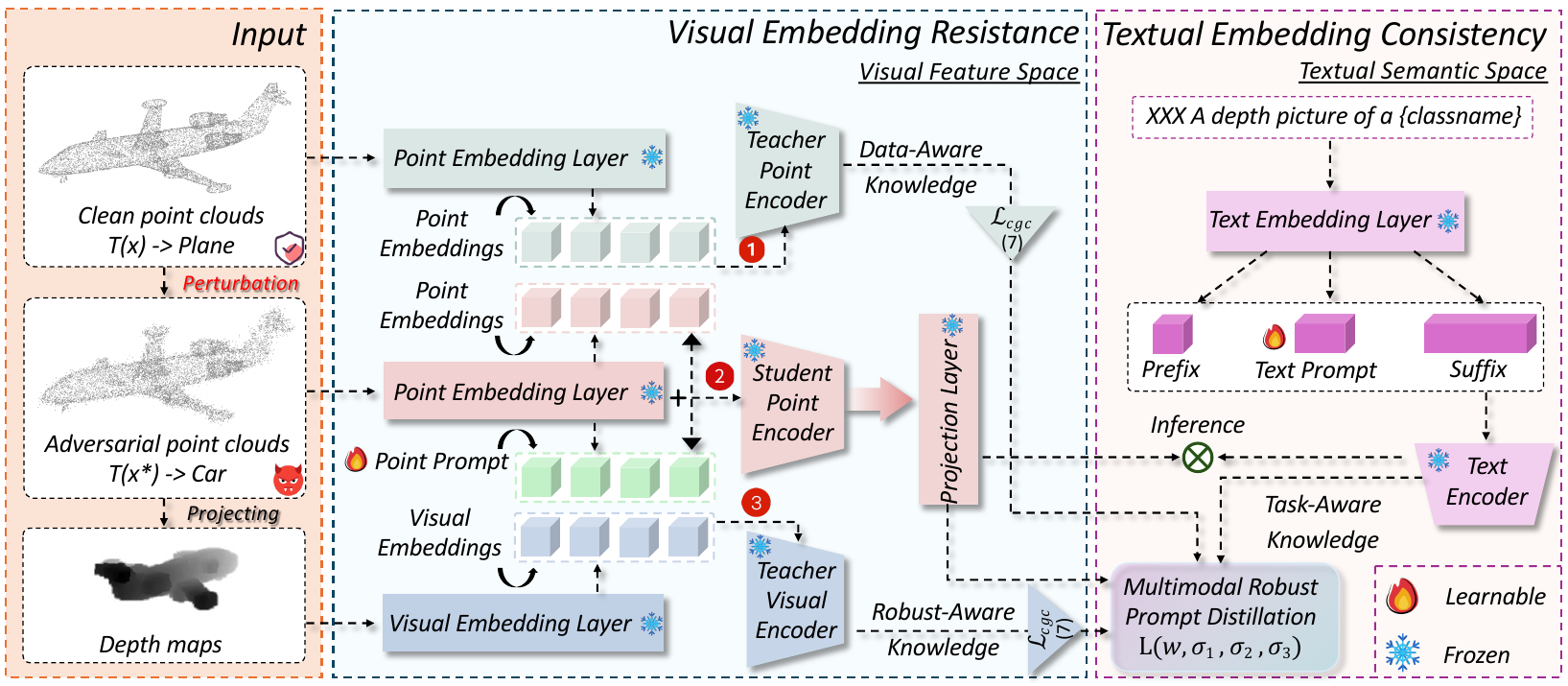}
    \caption{The proposed Multimodal Robust Prompt Distillation (MRPD) framework. During training, robust knowledge from three teachers (image, text, and a 3D model) is distilled into lightweight prompts. At inference, these prompts enhance the student model's robustness with zero additional computational cost.}
    \vspace{-8pt}
\label{fig:pipeline}
\end{figure*}
As shown in Figure~\ref{fig:pipeline}, we propose a framework to enhance the adversarial robustness of VLM-guided point cloud models via parameter-efficient multimodal prompt tuning. 
The key idea is to distill multimodal knowledge from image, text, and point cloud modalities into lightweight, learnable prompts within the point cloud student and its text encoder.
We first introduce our \textbf{MultiModal Robust Knowledge Collector}, which sources stable supervision signals from each modality. Subsequently, we describe the \textbf{Multi-Prompt Distillation} architecture, where these signals guide the optimization of the student model's prompts.

\subsection{MultiModal Knowledge Collector}

Enhancing the robustness of VLM-guided 3D models is a formidable challenge. Unlike in the 2D domain, the field of 3D deep learning lacks a wealth of powerful, pre-trained robust models that can serve as a foundation. Meanwhile, point cloud data is inherently vulnerable due to its sparsity, unstructured nature, and sensitivity to geometric perturbations. These characteristics make standard robust training techniques, such as adversarial training, not only computationally expensive but also prone to failure, as they rely on large-scale, high-quality data which is scarce for 3D shapes.
Given these domain-specific limitations, sourcing robust knowledge solely from the 3D modality is insufficient. We therefore propose to collect and aggregate multimodal supervision signals from three complementary modalities: the 2D projection domain, the semantic text domain, and 3D point cloud domain.

\subsubsection{Image Modality: A Source of Cross-Domain Stability}
We observe a helpful domain gap between 3D point clouds and their 2D image-space representations. \textit{Adversarial attacks meticulously crafted to perturb a 3D model's prediction often fail to transfer effectively when the point cloud is projected into a 2D image.} The geometric perturbations become subtle textural changes or are lost entirely during rendering. We leverage this phenomenon by using the frozen image encoder of the VLM, $f_I$, as a stable source of cross-modal robust knowledge. 

Specifically, for a given 3D point cloud $P$, we first render it into a multi-view depth image set via a projection operator $\Pi(\cdot)$. This process generates a view-invariant representation. The resulting image is then fed into the image encoder to extract a robust feature vector, $\mathbf{z}_I$:
\begin{equation}
    \mathbf{z}_I = \text{normalize}(f_I(\Pi(P)))
\end{equation}
This feature $\mathbf{z}_I$ remains remarkably stable even when $P$ is adversarially attacked, providing a consistent supervision signal for the student model.

\subsubsection{Text Modality: Learnable Prompts for Semantic Priors}
We introduce a learnable textual prompt to capture robust, task-specific semantic priors. Instead of relying on static, hand-crafted templates (\textit{e.g.}, ``a photo of a \{class\}''), we design a prompt that can be optimized to absorb semantic cues for our multimodal distillation process. 

Let $T = \{t_1, t_2, ..., t_N\}$ be the names of the downstream classes. We define a set of learnable prompt vectors $V_{ctx} = \{\mathbf{v}_1, ..., \mathbf{v}_M\}$. These vectors are concatenated with each class name to form a dynamic prompt:
\begin{equation}
T'_{cls} = \text{concat}(V_{ctx}, t_{cls})
\end{equation}
The complete set of dynamic prompts is then passed through the text encoder $f_T$ to produce robust class-level semantic embeddings $\mathbf{z}_T$:
\begin{equation}
\label{ZT}
    \mathbf{z}_T = \text{normalize}(f_T(\{T'_1, T'_2, ..., T'_N\}))
\end{equation}
By making the prompt context $V_{ctx}$ learnable, our model can adaptively find a semantic space that is more resilient to adversarial shifts.

\subsubsection{Point Cloud Modality: Structural Guidance from a Teacher}
While the 3D modality is vulnerable, a stronger, more capable point cloud model can still provide valuable geometric and structural knowledge. We employ a pre-trained point cloud model, denoted as the ``teacher'' encoder $f_P^{\text{teacher}}$, which remains frozen throughout training. This teacher model's role is to provide a consistent and high-quality representation of the original, \textit{clean} 3D shape.

For a clean point cloud $P$, the teacher produces a stable reference feature $\mathbf{z}_P^{\text{teacher}}$:
\begin{equation}
    \mathbf{z}_P^{\text{teacher}} = \text{normalize}(f_P^{\text{teacher}}(P))
\end{equation}
This feature serves as a geometric anchor, guiding the student model to maintain structural integrity even when its input is perturbed.
Collectively, these three components provide a rich and diverse set of robust features, $\mathbf{z}_I$, $\mathbf{z}_T$, and $\mathbf{z}_P^{\text{teacher}}$, that form the foundation of our MRPD framework. These supervision signals are then utilized to optimize the student model's learnable prompt. 

\subsection{Multimodal Robust Prompt Distillation}
Having established the collection of multimodal knowledge from image, text, and point cloud modalities, we now detail the core of our framework: the distillation architecture. Our goal is to transfer this multi-source knowledge into a lightweight ``student'' point cloud model, $f_P^{\text{student}}$, by optimizing a set of learnable prompts integrated within it. This process is designed to harden the student model against adversarial attacks without altering its backbone, ensuring zero overhead at inference. The distillation is governed by two key principles. First, to mitigate potential semantic conflicts and prevent the student from learning wrong teacher predictions, we introduce a \textbf{Confidence-Gated Distillation} loss. Second, to optimally balance the influence of the three diverse modalities, we employ a \textbf{Dynamic Weighting} mechanism that learns the relative importance of each teacher signal during training.

\subsubsection{Confidence-Gated Distillation for Reliable Knowledge Transfer}
A key challenge in multi-teacher distillation is handling cases where teachers disagree or are incorrect, especially on ambiguous or out-of-distribution samples. Blindly forcing the student to mimic all teachers can propagate errors and degrade performance. To address this, we propose a \textbf{Confidence-Gated Contrastive Loss}. This loss function acts as a quality filter, ensuring that knowledge is only distilled from a teacher when it is both correct and confident in its prediction.

Let $\mathbf{z}_{\text{stu}} \in \mathbb{R}^{B \times D}$ be the feature outputs from the student model for a batch of $B$ inputs, and $\mathbf{z}_{\text{ref}} \in \mathbb{R}^{B \times D}$ be the corresponding features from one of the teacher modalities (image, text, or point cloud). Let $\mathbf{z}_T \in \mathbb{R}^{C \times D}$ be the set of $C$ class-level text embeddings from (\ref{ZT}), which serve as the semantic ground truth for classification.
First, we assess the teacher's prediction confidence by computing the logits between its features and the class embeddings:
\begin{equation}
    \mathcal{L}_{\text{ref}} = \frac{\mathbf{z}_{\text{ref}} \cdot \mathbf{z}_T^\top}{\tau}
    \label{eq:ref_logits}
\end{equation}
where $\tau$ is a temperature hyperparameter.
Next, for each sample $i$ in the batch with ground-truth label $y_i$, we create a binary mask $\mathcal{M}$ that validates the teacher's prediction. A sample is considered valid for distillation only if its true label is within the top-$k$ predictions made by the teacher:
\begin{equation}
    \mathcal{M}_i = 
    \begin{cases}
    1 & \text{if } y_i \in \text{top-k}(\mathcal{L}_{\text{ref}, i}) \\
    0 & \text{otherwise}
    \end{cases}
    \label{eq:mask}
\end{equation}
This mask $\mathcal{M}$ effectively filters out instances where the teacher is likely incorrect, preventing negative knowledge transfer.
Finally, we compute a symmetric contrastive loss only on the subset of samples where $\mathcal{M}_i = 1$. Let $\mathbf{z}_{\text{stu}}^{\text{sel}}$ and $\mathbf{z}_{\text{ref}}^{\text{sel}}$ be the student and reference features selected by the mask. The Confidence-Gated Contrastive Loss, $\mathcal{L}_{\text{CGC}}$ is:
\begin{equation}
\mathcal{L}_{\text{CGC}}(\mathbf{z}_{\text{stu}}, \mathbf{z}_{\text{ref}}) = \frac{1}{2} \left( \text{CE}(\mathbf{S}, \mathbf{y}) + \text{CE}(\mathbf{S}^\top, \mathbf{y}) \right)
\label{eq:cgc_loss}
\end{equation}
where $\mathbf{S} = (\mathbf{z}_{\text{stu}}^{\text{sel}} \cdot (\mathbf{z}_{\text{ref}}^{\text{sel}})^\top) / \tau$ is the similarity matrix between the selected student and reference features, $\mathbf{y}$ is a vector of identity labels $[0, 1, \dots, B_{sel}-1]$, and CE denotes the Cross-Entropy loss. This loss pulls the student's representation of a valid sample towards its teacher's representation, while pushing it away from other samples in the batch.

\subsubsection{Dynamic Loss Weighting for Multimodal Balancing}
The three teacher modalities provide complementary but distinct types of knowledge. The image teacher offers holistic visual cues, the text teacher provides semantic priors, and the point cloud teacher gives fine-grained geometric guidance. Statically assigning weights to their respective distillation losses is suboptimal, as their relative importance may vary across different data samples and training stages.
To this end, we employ a dynamic weighting strategy inspired by multi-task learning~\cite{kendall2018multi} to automatically balance their contributions. This method frames the total loss as a multi-task objective where each modality's loss has an associated learnable uncertainty parameter. The model learns to down-weigh modalities with higher uncertainty (\textit{i.e.}, less reliable signals).

Let $\mathcal{L}_{I}$, $\mathcal{L}_{P}$, and $\mathcal{L}_{T}$ be the Confidence-Gated Contrastive Losses calculated using the image, point cloud, and text teacher features, respectively, as defined in Eq.~\ref{eq:cgc_loss}. We introduce three learnable log-variance parameters, $\lambda_I, \lambda_P, \lambda_T$. The final, combined distillation loss $\mathcal{L}_{\text{total}}$ is formulated as:

\begin{equation}
\label{eq:final_loss} 
\mathcal{L}_{\text{total}} = \sum_{k \in \{I, P, T\}} \left( e^{-\lambda_k} \mathcal{L}_{k} + \lambda_k \right)
\end{equation}
In this formulation, each $\exp(-\lambda_k) \mathcal{L}_k$ term is an uncertainty-weighted loss, and $\lambda_k$ acts as a regularizer to prevent the weights from growing infinitely. This allows the model to learn an optimal, data-driven balance between the three robust supervision signals, leading to more stable and effective distillation.

\subsection{Training and Evaluation}

Our framework consists of two phases. \textbf{During training}, we perform Multimodal Robust Prompt Distillation. Specifically, we optimize a set of lightweight prompts for the student's point cloud encoder and the text encoder. This process is guided by a dynamic, confidence-gated distillation loss that aggregates supervision signals from image, text, and 3D teacher models.
\textbf{At inference}, the entire distillation apparatus (teachers and loss mechanism) is discarded. The student model operates solely with its optimized prompts, achieving enhanced robustness with \textbf{no additional computational overhead or architectural changes}. This efficiency makes our method highly practical for deployment. The full procedure is outlined in \textit{Algorithm 1 in supplementary materials}.

\begin{table*}[htbp]
\centering
\small 
\sisetup{detect-weight, mode=text, table-format=2.2, table-space-text-post=\%, table-number-alignment=center}
\setlength{\tabcolsep}{5pt} 
\begin{tabular}{
  l
  S[table-format=2.2]
  S[table-format=2.2] S[table-format=2.2] S[table-format=2.2] S[table-format=2.2] S[table-format=2.2] S[table-format=2.2] S[table-format=2.2] S[table-format=2.2]
  S[table-format=2.2] 
  S[table-format=7.0]
}
\toprule
\textbf{Method} & {\textbf{Clean}} & {\textbf{PGD}} & {\textbf{Perturb}} & {\textbf{KNN}} & {\textbf{ADD-CD}} & {\textbf{ADD-HD}} & {\textbf{AOF}} & {\textbf{Drop-200}}& {\textbf{AdvPC}}& {\textbf{Avg. R}} & {\textbf{+Params}}  \\
\midrule

\multicolumn{12}{l}{\textit{\textbf{Dataset: ModelNet40}}} \\
\addlinespace[0.3em]
Clean Model     & 70.99 & 63.82 &  0.00          & 51.86          &  0.04          &  0.00          &  0.00          & 53.04          &  0.00          & 21.10          & 0 \\
SRS             & 55.75 & 54.90 & 44.12          & 53.00          & 61.83          & 46.11          & 13.25          & 31.36          & 16.57          & 40.14          & 0 \\
SOR             & 65.56 & 61.95 & 50.12          & 59.20          & 60.86          & 55.06          & 17.26          & 41.82          & 19.81          & 45.76          & 0 \\
DUP             & 61.67 & 61.06 & 52.35          & 62.44          & 54.54          & 43.96          & 25.61          & 39.30          & 20.54          & 44.98          & 814307 \\
IF-Defense      & 62.93 & 62.40 & 63.37          & 62.80          & 63.09          & 62.40          & 50.36          & 63.57          & \textbf{47.65} & 59.46          & 1978209 \\
Adv Training    & 89.95 & 88.01 & 68.15          & 85.70          & 68.60          & 56.81          & 54.74          & 72.45          & 32.62          & 65.89          & 40960 \\
\rowcolor{gray!15}
\textbf{MRPD (Ours)} & \textbf{90.52} & \textbf{89.14} & \textbf{80.79} & \textbf{87.88} & \textbf{81.36} & \textbf{68.31} & \textbf{54.86} & \textbf{78.32} & 39.99          & \textbf{72.58} & 0 \\
\midrule

\multicolumn{12}{l}{\textit{\textbf{Dataset: ScanObjectNN}}} \\
\addlinespace[0.3em]
Clean Model     & 52.74          & 25.99          &  0.00          &  4.86          &  0.00          &  0.00          &  0.00          & 49.03          &  0.00          & 9.99           & 0 \\
SRS             & 51.73          & 41.81          & 39.14          & 35.88          & 43.30          & 20.12          &  7.43          & 46.63          & 14.54          & 31.11          & 0 \\
SOR             & 45.98          & 38.83          & 32.58          & 33.17          & 37.30          & 31.71          & 20.16          & 41.50          & 21.17          & 32.05          & 0 \\
DUP             & 32.69          & 31.99          & 29.49          & 31.71          & 31.68          & 28.42          & 17.14          & 27.55          & 17.35          & 26.92          & 814307 \\
IF-Defense      & 39.87          & 39.73          & 39.83          & 40.49          & 39.73          & 40.60          & 34.80          & 36.71          & 33.62          & 38.19          & 1978209 \\
Adv Training    & \textbf{83.48} & \textbf{79.15} & 69.81          & 63.43          & 74.15          & 48.82          & 57.29          & \textbf{80.40} & 40.39          & 64.18          & 15360 \\
\rowcolor{gray!15}
\textbf{MRPD (Ours)} & 78.80 & 76.13 & \textbf{72.24} & \textbf{72.52} & \textbf{74.74} & \textbf{57.08} & \textbf{60.06} & 77.72 & \textbf{48.65} & \textbf{67.39} & 0 \\
\bottomrule
\end{tabular}
\caption{Classification accuracy (\%) of different defense strategies under various \textbf{white-box attacks} on ModelNet40 and ScanObjectNN. Best results in each column are in \textbf{bold}. \textbf{Avg. R} means average robust accuracy of all attacks listed.}
\label{table:white-box}
\end{table*}

\section{Experiments}
\subsection{Experimental Setup}
\noindent\textbf{Datasets and Metrics.}
We evaluate our method on two standard benchmarks: the synthetic \textbf{ModelNet40}~\cite{wu20153d} (40 classes, 1024 points/object) and the real-world \textbf{ScanObjectNN}~\cite{uy2019revisiting} (15 classes, with background noise and occlusions). We report classification accuracy (\%) on both clean data (\textbf{Clean Accuracy}) and adversarially perturbed data (\textbf{Robust Accuracy}) to measure performance and resilience.
\noindent\textbf{Attack Scenarios.}
We test our defense against a comprehensive suite of attacks under both \textbf{white-box} (full model access) and \textbf{black-box} (transfer-based) settings. The attacks include point-wise perturbations (\textbf{PGD}~\cite{liu2019extending}, \textbf{Perturb}, \textbf{KNN}~\cite{xiao2018generating}), point additions (\textbf{ADD-CD/HD}~\cite{wen2020geometry}, \textbf{AOF}~\cite{zhang2023art}), \textbf{AdvPC}~\cite{hamdi2020advpc}, and point removal (\textbf{Drop-200}). This diverse set of threats allows for a rigorous evaluation of our model's robustness.

\subsection{Main Experimental Results}
\noindent\textbf{White-Box Attack Analysis.}
Table~\ref{table:white-box} showcases the white-box robustness of our MRPD framework. The results confirm its ability to achieve superior defense across a broad spectrum of attacks while introducing zero inference overhead.
On ModelNet40, MRPD establishes a new state-of-the-art, achieving the highest average robustness (72.58\%) and even surpassing the strong adversarial training baseline on clean data (90.52\% vs. 89.95\%). Its significant gains against diverse attacks like Perturb (+12.64\%) and ADD-HD (+11.50\%) highlight that our multimodal distillation learns a more generalizable feature space, avoiding the overfitting common to adversarial training.
This advantage extends to the challenging real-world ScanObjectNN dataset, where MRPD again delivers a higher average robust accuracy (67.39\% vs. 64.18\%). Its strong performance on noisy and occluded data validates our core premise: distilling knowledge from stable 2D and text teachers enables the model to preserve essential features against both adversarial attacks and real-world imperfections. Ultimately, by embedding multimodal robustness into lightweight prompts, MRPD offers a powerful and practical defense that breaks the trade-off between security and efficiency.

\noindent\textbf{Generalization Against Black-Box Attacks.}
We further assess MRPD's generalizability in a challenging black-box setting, where attacks are transferred from a known, third-party model. As shown in Table~\ref{table:blackbox}, our method demonstrates remarkable resilience, confirming its robustness extends beyond specific threat models.
On ModelNet40, MRPD again achieves superior performance with the highest average robust accuracy (65.72\%). It significantly outperforms the adversarial training baseline on a majority of transferable attacks, including ADD-CD (+9.16\%) and ADD-HD (+9.93\%). This strong generalization suggests that by learning from diverse 2D, 3D, and text teachers, MRPD develops a decision boundary less correlated with standard 3D architectures, making it inherently more resistant to transferred attacks.
This pattern is amplified on the real-world ScanObjectNN dataset, where MRPD once again secures the highest average robustness (67.47\%). Its notable performance across a wide spectrum of threats, especially against challenging attacks like AOF and AdvPC, shows that the distilled multimodal knowledge provides resilience against both transferred perturbations and the inherent noise of real-world data. This black-box evaluation confirms that MRPD's multimodal distillation paradigm yields a fundamentally more generalizable defense.

\begin{table*}[htbp]
\centering
\small 
\sisetup{detect-weight, mode=text, table-format=2.2, table-space-text-post=\%, table-number-alignment=center}
\setlength{\tabcolsep}{5pt} 

\begin{tabular}{
  l
  S[table-format=2.2]
  S[table-format=2.2] S[table-format=2.2] S[table-format=2.2] S[table-format=2.2] S[table-format=2.2] S[table-format=2.2] S[table-format=2.2] S[table-format=2.2]
  S[table-format=2.2] 
  S[table-format=7.0]
}
\toprule
\textbf{Method} & {\textbf{Clean}} & {\textbf{PGD}} & {\textbf{Perturb}} & {\textbf{KNN}} & {\textbf{ADD-CD}} & {\textbf{ADD-HD}} & {\textbf{AOF}} & {\textbf{Drop-200}}& {\textbf{AdvPC}}& {\textbf{Avg. R}} & {\textbf{+Params}} \\
\midrule

\multicolumn{11}{l}{\textit{\textbf{Dataset: ModelNet40}}} \\
\addlinespace[0.3em]
Clean Model     & 70.99 & 69.37 & 52.63 & 58.39 & 57.37 & 41.49 & 20.62 & 56.65 & 18.64 & 46.90 & 0 \\
SRS             & 55.75 & 54.66 & 42.54 & 47.93 & 51.78 & 37.52 & 17.38 & 33.55 & 16.98 & 37.79 & 0 \\
SOR             & 65.56 & 62.76 & 48.99 & 54.01 & 59.93 & 51.99 & 20.18 & 43.07 & 18.40 & 44.92 & 814307 \\
DUP             & 61.67 & 61.06 & 45.66 & 53.61 & 46.43 & 39.91 & 16.73 & 41.73 & 15.76 & 40.11 & 0 \\
IF-Defense      & 62.93 & 62.76 & 61.59 & 61.55 & 58.43 & 50.28 & \bfseries 40.80 & 52.96 & \bfseries 38.09 & 53.31 & 1978209 \\
Adv Training    & 89.95 & 89.42 & 74.03 & 82.13 & 63.94 & 44.69 & 38.09 & 77.67 & 29.34 & 62.41 & 40960 \\
\rowcolor{gray!15}
\textbf{MRPD (Ours)} & \textbf{90.52} & \textbf{89.55 }& \textbf{77.84} & \textbf{84.52} & \textbf{73.10} & \textbf{54.62} & 36.22 & \textbf{83.51} & 26.42 & \bfseries 65.72 & 0\\
\midrule

\multicolumn{11}{l}{\textit{\textbf{Dataset: ScanObjectNN}}} \\
\addlinespace[0.3em]
Clean Model     & 52.74 & 50.90 & 37.82 & 40.42 & 39.63 & 30.85 & 15.13 & 47.88 & 14.26 & 34.61 & 0 \\
SRS             & 51.73 & 49.62 & 37.44 & 41.43 & 39.52 & 30.50 & 15.27 & 46.15 & 14.23 & 34.27 & 0 \\
SOR             & 45.98 & 44.62 & 34.00 & 37.23 & 39.49 & 37.99 & 20.89 & 41.57 & 20.51 & 34.54 & 814307 \\
DUP             & 32.69 & 32.96 & 29.63 & 31.37 & 30.36 & 29.01 & 16.48 & 28.14 & 16.83 & 26.85 & 0 \\
IF-Defense      & 39.87 & 40.28 & 40.56 & 40.46 & 39.63 & 38.65 & 32.34 & 37.13 & 33.52 & 37.82 & 1978209 \\
Adv Training    & \bfseries 83.48 & \bfseries 82.55 & 74.05 & 73.46 & 71.24 & 61.55 & 40.18 & \bfseries 81.44 & 41.57 & 65.76 & 15360 \\
\rowcolor{gray!15}
\textbf{MRPD (Ours)} & 78.80 & 78.87 & \textbf{75.16 }& \textbf{75.09} & \textbf{74.81} & \textbf{67.28} & \textbf{44.07} & 77.93 &\textbf{ 46.56} & \bfseries 67.47 & 0 \\

\bottomrule
\end{tabular}
\caption{Classification accuracy (\%) of different defense strategies under various \textbf{blackbox attacks} on ModelNet40 and ScanObjectNN. Best results in each column are in \textbf{bold}. \textbf{Avg. R} means average robust accuracy of all attacks listed.}
\label{table:blackbox}
\end{table*}

\noindent\textbf{Qualitative Analysis.}
\begin{figure}[t]
    \centering
    \includegraphics[width=1.0\linewidth]{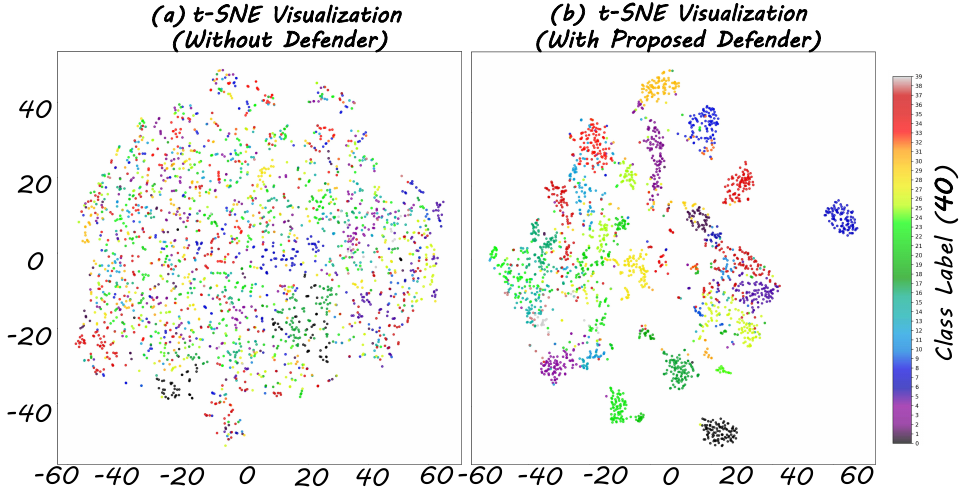}
    \caption{MRPD preserves feature space integrity under adversarial attack on ModelNet40. (a) Without defense, features from different classes become indistinguishable. (b) With MRPD, features remain well-separated, ensuring robust classification.}
    \vspace{-8pt}
    \label{fig:tsne}
\end{figure}
To provide a qualitative understanding of our method's effectiveness, we visualize the feature embeddings of adversarial examples from ModelNet40 using t-SNE~\cite{van2008visualizing}, as shown in Figure~\ref{fig:tsne}.
The visualization reveals a stark contrast. Without defense (Figure~\ref{fig:tsne}(a)), adversarial attacks cause the feature space to collapse, resulting in chaotic and overlapping clusters where class distinctions are lost. This leads to widespread misclassification. In sharp contrast, the feature space of our MRPD-protected model (Figure~\ref{fig:tsne}(b)) remains highly structured even under the same attacks. The features form compact, well-separated clusters corresponding to their true classes. This demonstrates that our method successfully preserves the semantic structure of the feature space, ensuring robust class separability and correct classification against adversarial manipulations. Similar improvements are observed on the more challenging real-world ScanObjectNN dataset, as detailed in \textit{Figure 1 in the supplementary materials}.

\subsection{Ablation Studies}
To validate the contributions of each component in our MRPD framework, we conduct a series of ablation studies. We analyze the effectiveness of the multimodal prompts, the dynamic loss weighting, the confidence-gated distillation loss, and the prompt parameterization. For brevity, we present the key results here, while a more comprehensive breakdown of results across all attack types and additional analyses are provided in the \textit{supplementary materials}.

\subsection{Effectiveness of MultiModal Prompts}
\begin{table}[t]
\centering
\small 
\sisetup{detect-weight, mode=text, table-format=2.2, table-number-alignment=center}
\setlength{\tabcolsep}{4pt} 

\begin{tabular}{l S S S S S}
\toprule
\textbf{Method} & {\textbf{Clean}} & {\textbf{PGD}} & {\textbf{ADD-CD}} & {\textbf{Drop}} & {\textbf{Avg. R}} \\
\midrule
\multicolumn{6}{l}{\textit{\textbf{Dataset: ModelNet40}}} \\
\addlinespace[0.3em]
Baseline      & 70.99 & 63.82 & 0.04  & 53.04 & 21.10 \\
+ Text Prompt & 86.18 & 80.83 & 52.67 & 70.54 & 52.37 \\
+ Point Prompt& 75.89 & 73.82 & 67.95 & 61.59 & 55.16 \\
\rowcolor{gray!15}
\textbf{MRPD (Full)}   & \bfseries 90.52 & \bfseries 89.14 & \bfseries 81.36 & \bfseries 78.32 & \bfseries 72.58 \\
\midrule
\multicolumn{6}{l}{\textit{\textbf{Dataset: ScanObjectNN}}} \\
\addlinespace[0.3em]
Baseline      & 52.74 & 25.99 & 0.00  & 49.03 & 9.99 \\
+ Text Prompt & 69.26 & 55.24 & 49.41 & 64.26 & 44.56 \\
+ Point Prompt& 53.37 & 46.01 & 45.80 & 49.48 & 35.67 \\
\rowcolor{gray!15}
\textbf{MRPD (Full)}   & \bfseries 78.80 & \bfseries 76.13 & \bfseries 74.74 & \bfseries 77.72 & \bfseries 67.39 \\
\bottomrule
\end{tabular}
\caption{Ablation study of MRPD components on ModelNet40 and ScanObjectNN. We show accuracy (\%) on clean data, three representative attacks, and the recalculated average robustness. \textbf{Avg. R} denotes the average accuracy over all white-box attacks.}
\label{tab:ablation_white-box_condensed}
\end{table}
To isolate the contributions of our multimodal prompts, we compare our full MRPD model against a Baseline (no prompts) and single-prompt variants in Table~\ref{tab:ablation_white-box_condensed}. The Baseline is extremely vulnerable, with accuracy collapsing on attacks like `ADD-CD', highlighting the need for defense. Introducing either the text or point prompt individually yields significant gains in both clean and robust accuracy, confirming their value in providing semantic or geometric resilience. Crucially, our full MRPD model, integrating both prompt types, substantially outperforms all other configurations. Its average robust accuracy on ModelNet40 (72.58\%) and ScanObjectNN (67.39\%) demonstrates a massive improvement over single-prompt variants. This performance leap, far exceeding the sum of individual contributions, confirms a powerful synergistic effect. Fusing multimodal knowledge into both point and text encoders via our distillation is therefore essential for achieving a comprehensive, sota defense.

\subsection{Analysis of Dynamic Loss Weighting}
\begin{figure}[h!]
  \centering
  \includegraphics[width=1\columnwidth]{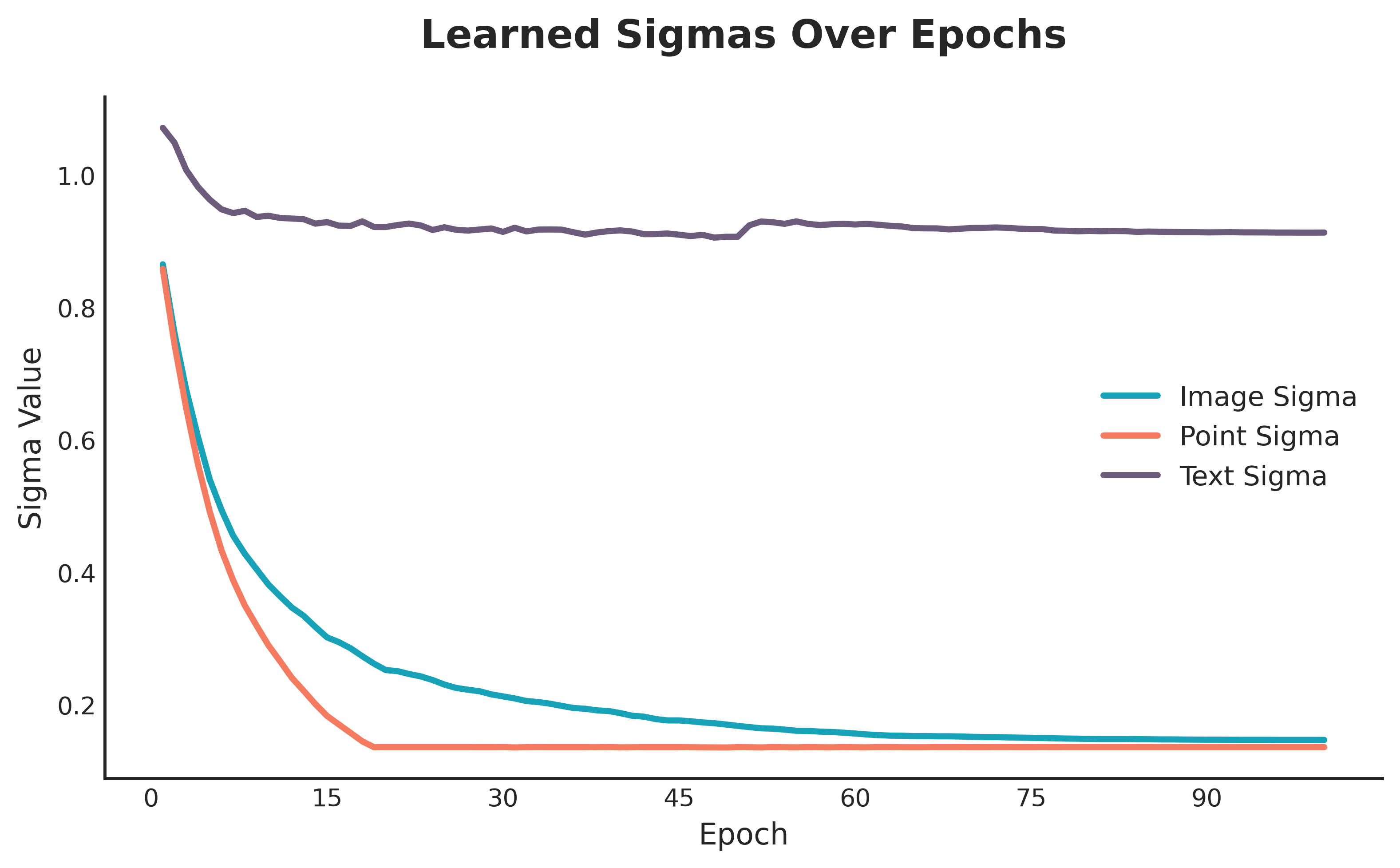} 
  \caption{\textbf{Evolution of learned loss weights ($1/\sigma^2$) for each distillation task.} The model learns to heavily prioritize the point and image teachers while using the text teacher as a low-weight semantic regularizer.}
  \label{fig:loss_weights}
\end{figure}

To validate our dynamic loss weighting, we analyzed the evolution of the learned weights ($w_k = \exp(-\lambda_k)$) during training on ModelNet40. As shown in Figure~\ref{fig:loss_weights}, the model automatically learns a sophisticated balancing strategy. The weights for the point and image teachers increase dramatically in early epochs, stabilizing at high values ($w_P \approx 55, w_I \approx 47$), while the text teacher's weight ($w_T$) remains consistently low ($\approx 1.2$). This learned hierarchy reveals that the model prioritizes rich geometric guidance from the point and image teachers to build a robust feature foundation, while treating the text teacher as a high-level semantic regularizer. This emergent behavior confirms the efficacy of our dynamic scheme in achieving a more effective and stable balance than manually fixed weights could provide.

\subsection{Ablation on Confidence-Gated Distillation Loss}
\begin{table}[h]
\centering
\sisetup{detect-weight, mode=text, table-format=2.2, table-number-alignment=center}
\setlength{\tabcolsep}{4pt} 
\begin{tabular}{
l
S[table-format=2.2]
S[table-format=2.2]
S[table-format=2.2]
S[table-format=2.2]
S[table-format=2.2, detect-weight]
}
\toprule
\textbf{Method} & {\textbf{Clean}} & {\textbf{PGD}} & {\textbf{Perturb}} & {\textbf{AOF}} & {\textbf{Avg. R}} \\
\midrule
w/o CGC & 90.40 & \bfseries 89.47 & 79.21 & 52.88 & 70.99 \\
\rowcolor{gray!15}
\textbf{MRPD (Ours)} & \textbf{90.52} & 89.14 & \textbf{80.79} & \textbf{54.86} & \bfseries 72.58 \\
\bottomrule
\end{tabular}
\caption{Ablation of the Confidence-Gated Contrastive (CGC) loss on ModelNet40. \textbf{Avg. R} denotes the average accuracy over all white-box attacks. Best results are in \textbf{bold}.}
\label{tab:ablation_cgc_small}
\end{table}
\noindent We perform an ablation study on the Confidence-Gated Contrastive (CGC) loss to highlight its importance. We compare our full MRPD with a variant (w/o CGC) that uses a standard, ungated multi-teacher distillation.
As shown in Table~\ref{tab:ablation_cgc_small}, the w/o CGC variant performs marginally better on the standard PGD attack, likely due to its simple averaging of teacher knowledge. However, this naive approach falters on more complex attacks. Our full model with CGC shows superior performance on Perturb (+1.58\%) and AOF (+1.98\%), where CGC's ability to filter conflicting teacher signals in ambiguous cases is critical. This strategic gating is key to achieving higher overall robustness (Avg. R), confirming CGC as a vital component of our framework.

\subsection{Ablation on Prompt Parameters}

\begin{table}[h]
\centering
\sisetup{detect-weight, mode=text, table-format=2.2, table-number-alignment=center}
\setlength{\tabcolsep}{4pt} 
\begin{tabular}{cc S S S S S}
\toprule
\multicolumn{2}{c}{\textbf{Prompt Size}} & \multicolumn{5}{c}{\textbf{Accuracy (\%)}} \\
\cmidrule(lr){1-2} \cmidrule(lr){3-7}
\textbf{Point} & \textbf{Text} & \textbf{Clean} & \textbf{PGD} & \textbf{ADD-CD} & \textbf{Drop} & \textbf{Avg. R} \\
\midrule
5 & 3 & 89.71 & 87.93 & 78.97 & 78.97 & 69.91 \\
\rowcolor{gray!15}
\textbf{10} & \textbf{3} & \bfseries 90.52 & \bfseries 89.14 & \bfseries 81.36 & \bfseries 78.32 & \bfseries 72.58 \\
10 & 5 & 89.71 & 88.70 & 79.38 & 77.55 & 70.78 \\
15 & 3 & 89.38 & 88.41 & 79.70 & 76.74 & 69.94 \\
\bottomrule
\end{tabular}
\caption{Ablation on prompt size using ModelNet40. We report accuracy (\%) on clean data and representative white-box attacks. \textbf{Avg. R} denotes the average accuracy over all white-box attacks. The optimal configuration is highlighted.}
\label{tab:ablation_prompt_size_main}
\end{table}

\noindent To determine the optimal prompt configuration, we ablate the number of point and text tokens. As shown in Table~\ref{tab:ablation_prompt_size_main}, a configuration of 10 point tokens and 3 text tokens strikes the best balance, achieving the highest average robustness (72.58\%) without compromising clean accuracy. Increasing the point prompt size from 5 to 10 significantly boosts performance, but further increasing it to 15 leads to a decline, suggesting potential overfitting. Similarly, a concise 3-token text prompt proves most effective. This highlights the importance of prompt capacity: sufficient to capture robust knowledge, yet compact enough to avoid overfitting. We adopt the (10, 3) configuration for all experiments.

\section{Conclusion}
In this paper, we introduce Multimodal Robust Prompt Distillation, a novel framework that enhances 3D model robustness by distilling knowledge from image, text, and 3D teachers into lightweight prompts. The core of our MRPD framework is a novel distillation architecture that translates robust knowledge from powerful image, text, and 3D teachers into complementary point and text prompts. To manage the potential for conflicting advice arising from these distinct modalities, our Confidence-Gated Contrastive loss acts as a dynamic arbiter, selectively filtering inconsistent signals to stabilize the learning process and distill a truly robust representation. Experiments show MRPD significantly outperforms existing methods against a wide range of attacks, especially on real-world data, while critically adding zero inference overhead. This work establishes an effective and practical paradigm for robust 3D vision. Future directions include applying this prompt-distillation technique to other 3D tasks like detection and segmentation.

\section{Acknowledgments}
Xiang Gu, Liming Lu and Shuchao Pang are supported by the National Natural Science Foundation of China (Grant No.62206128), National Key Research and Development Program of China under (Grant No.2023YFB2703900) and the Postgraduate Research \& Practice Innovation Program of Jiangsu Province (Grant No.KYCX24\_0723).
Anan Du is supported by the Start-up Fund for New Talented Researchers
of Nanjing University of Industry Technology (Grant No.YK24-05-04).
Yongbin Zhou is supported by the National Natural Science Foundation of China (Grant No.U2336205).

\bibliography{aaai2026}
\clearpage 
\appendix        
\section*{Supplementary Material}

In these supplementary materials, we provide additional details and results to further support the claims made in the main paper. We begin in \textbf{Section 1} by establishing the preliminaries, where we define our key notations (Table~\ref{tab:notation}), provide background on VLM-guided models and adversarial attacks, and present the detailed pseudocode for our MRPD framework (Algorithm~\ref{alg:mmpd}). Following this, \textbf{Section 2} presents a comprehensive set of extended ablation studies. These studies dissect the contributions of our framework's core components, including detailed analyses of the multi-modal prompts (Table~\ref{tab:ablation_white-box_full}), distillation teachers (Table~\ref{tab:ablation_teacher_white-box_full}), and prompt parameters (Table~\ref{tab:ablation_prompt_size_full}), and a qualitative analysis on ScanObjectNN (Figure~\ref{fig:tsne_scanobjectnn}). Finally, \textbf{Section 3} covers the experiment details to ensure full reproducibility, detailing our implementation specifics and the methodology for generating black-box attack data, including performance metrics of the surrogate model (Table~\ref{tab:surrogate_performance}).

\section{Preliminaries}
\subsection{Notation}
\label{sec:appendix_notation}

To ensure clarity and reproducibility, we summarize the key mathematical notations used throughout the paper in Table~\ref{tab:notation}. This table provides a quick reference for the symbols representing the models, data, features, and key components of our proposed Multimodal Robust Prompt Distillation (MRPD) framework.
\begin{table}[h!]
\centering
\setlength{\tabcolsep}{5pt} 
\renewcommand{\arraystretch}{1.2} 
\begin{tabular}{ll}
\toprule
\textbf{Symbol} & \textbf{Description} \\
\midrule
\multicolumn{2}{l}{\textit{\textbf{Data and Models}}} \\
$P$ & An input 3D point cloud. \\
$y$ & Ground-truth label for an input point cloud. \\
$f_{\text{stu}}$ & The student point cloud encoder model. \\
$f_I, f_T, f_P$ & Frozen teacher encoders for image, text, and \\
& point cloud modalities, respectively. \\
$\Pi(\cdot)$ & Operator to project a 3D point cloud to a \\
& 2D multi-view depth image. \\
\addlinespace[0.3em]
\multicolumn{2}{l}{\textit{\textbf{Learnable Parameters}}} \\
$\theta_p$ & Learnable parameters of the point prompt. \\
$\theta_t$ & Learnable parameters of the text prompt ($V_{ctx}$). \\
$\lambda_I, \lambda_P, \lambda_T$ & Learnable parameters for dynamically \\
& weighting the distillation losses. \\
\addlinespace[0.3em]
\multicolumn{2}{l}{\textit{\textbf{Features and Embeddings}}} \\
$\mathbf{z}_{\text{stu}}$ & Feature embedding from the student model. \\
$\mathbf{z}_I, \mathbf{z}_P$ & Feature embeddings from the image and \\
& point cloud teachers. \\
$\mathbf{z}_T$ & Set of class-level semantic embeddings \\
& from the text teacher. \\
\addlinespace[0.3em]
\multicolumn{2}{l}{\textit{\textbf{Losses and Mechanisms}}} \\
$\mathcal{L}_{\text{CGC}}$ & The Confidence-Gated Contrastive loss. \\
$\mathcal{M}$ & Binary mask for the confidence-gating mechanism. \\
$\mathcal{L}_{\text{total}}$ & The final combined multi-modal distillation loss. \\
$\tau$ & Temperature hyperparameter for scaling logits. \\
$k$ & Top-k hyperparameter for the confidence gate. \\
\bottomrule
\end{tabular}
\caption{Summary of key notations used in this paper.}
\label{tab:notation}
\end{table}

\begin{table*}[h!]
\centering
\small
\sisetup{detect-weight, mode=text, table-format=2.2, table-number-alignment=center}
\setlength{\tabcolsep}{4.5pt}

\begin{tabular}{l S S S S S S S S S S}
\toprule
\textbf{Method} & {\textbf{Clean}} & {\textbf{PGD}} & {\textbf{Perturb}} & {\textbf{KNN}} & {\textbf{ADD-CD}} & {\textbf{ADD-HD}} & {\textbf{Drop-200}} & {\textbf{AdvPC}} & {\textbf{AOF}} & {\textbf{Avg. Robust}} \\
\midrule
\multicolumn{11}{l}{\textit{\textbf{Dataset: ModelNet40}}} \\
\addlinespace[0.3em]
Baseline (w/o Prompts) & 70.99 & 63.82 & 0.00  & 51.86 & 0.04  & 0.00  & 53.04 & 0.00  & 0.00  & 21.10 \\
+ Text Prompt          & 86.18 & 80.83 & 49.55 & 72.12 & 52.67 & 45.75 & 70.54 & 18.64 & 28.89 & 52.37 \\
+ Point Prompt         & 75.89 & 73.82 & 63.41 & 72.08 & 67.95 & 52.84 & 61.59 & 24.07 & 25.53 & 55.16 \\
\rowcolor{gray!15}
\textbf{MRPD (Full)}   & \bfseries 90.52 & \bfseries 89.14 & \bfseries 80.79 & \bfseries 87.88 & \bfseries 81.36 & \bfseries 68.31 & \bfseries 78.32 & \bfseries 39.99 & \bfseries 54.86 & \bfseries 72.58 \\
\midrule
\multicolumn{11}{l}{\textit{\textbf{Dataset: ScanObjectNN}}} \\
\addlinespace[0.3em]
Baseline (w/o Prompts) & 52.74 & 25.99 & 0.00  & 4.86  & 0.00  & 0.00  & 49.03 & 0.00  & 0.00  & 9.99 \\
+ Text Prompt          & 69.26 & 55.24 & 44.62 & 27.65 & 49.41 & 39.90 & 64.26 & 30.81 & 44.59 & 44.56 \\
+ Point Prompt         & 53.37 & 46.01 & 43.10 & 40.25 & 45.80 & 27.06 & 49.48 & 18.53 & 15.09 & 35.67 \\
\rowcolor{gray!15}
\textbf{MRPD (Full)}   & \bfseries 78.80 & \bfseries 76.13 & \bfseries 72.24 & \bfseries 72.52 & \bfseries 74.74 & \bfseries 57.08 & \bfseries 77.72 & \bfseries 48.65 & \bfseries 60.06 & \bfseries 67.39 \\
\bottomrule
\end{tabular}
\caption{Detailed ablation study of MRPD components under all white-box attacks. ``Avg. Robust'' is the average accuracy across all 8 listed attacks. The best results are in \textbf{bold}.}
\label{tab:ablation_white-box_full}
\end{table*}

\subsection{VLM-guided Point Cloud Models}
Vision-Language Model (VLM) guided point cloud models represent a paradigm shift in 3D understanding, leveraging the rich semantic knowledge from pre-trained VLMs like CLIP. During their training phase, these encoders are taught to align features from three different modalities. Using large-scale text-image-point cloud triplet datasets, a point cloud encoder is trained to map 3D shapes into a joint embedding space that is already semantically structured by the powerful, pre-trained text and image encoders of the VLM. This process enables the point cloud model to inherit the VLM's robust semantic comprehension, allowing it to associate complex 3D geometry with natural language descriptions.

For a downstream task such as classification, the inference process operates as follows. First, a set of textual prompts describing the target classes, $T = \{t_1, t_2, ..., t_N\}$, is passed through the frozen text encoder $f_T$ to create a semantic classifier. The resulting text features are normalized to form a text embedding matrix $\mathbf{z}_T$:
\begin{equation}
    \mathbf{z}_T = \text{normalize}(f_T(T))
\end{equation}
Next, a given point cloud $P$ is fed into the point cloud encoder $f_P$ to generate its corresponding normalized feature vector $\mathbf{z}_P$:
\begin{equation}
    \mathbf{z}_P = \text{normalize}(f_P(P))
\end{equation}
The final prediction, $\hat{y}$, is determined by calculating the cosine similarity between the point cloud feature and all class-level text features, selecting the class with the highest correspondence:
\begin{equation}
    \hat{y} = \underset{i \in \{1,...,N\}}{\text{argmax}} \left( \mathbf{z}_P \cdot \mathbf{z}_{T_i}^\top \right)
\end{equation}

This architecture fundamentally differs from traditional models that are trained from scratch for specific tasks. While powerful, enhancing the robustness of these large VLM-guided models poses a significant challenge. Conventional robustness-enhancing techniques, such as adversarial training, are computationally expensive and data-hungry. Applying them directly to these large models can lead to prohibitive training costs and a high risk of overfitting, particularly in the context of point clouds where large, diverse, and clean datasets are scarce. Consequently, there is a pressing need for an efficient and parameter-light fine-tuning methodology to instill robustness without compromising the model's pre-trained knowledge or incurring massive computational overhead.

\subsection{Adversarial Attacks on Point Clouds.} 
Research on 3D adversarial attacks has rapidly evolved, initially drawing inspiration from 2D attacks. 
Early gradient-based methods, such as 3DAdv\cite{xiao2018generating}, adapted traditional attack pipelines using 3D-specific loss functions like the Chamfer distance. 
To enhance the stealthiness of these attacks, subsequent improvements like 3D-PGD\cite{liu2019extending} and GeoA3\cite{wen2020geometry} went beyond simple distance metrics, focusing on generating more realistic and less perceptible perturbations by considering geometric properties like tangent planes and local curvature. Recognizing the practical limitations of white-box attacks, which require full model access, another line of research has focused on the more challenging black-box scenarios. 
This includes heuristic approaches like EvolutionAdv\cite{cao2019adversarial}, which leverages evolutionary strategies, and query-based methods like \cite{wicker2019robustness}, which iteratively identifies and removes critical points. Beyond point-wise perturbation, researchers have also explored entirely different paradigms, such as using GANs to produce adversarial point clouds (AdvPC\cite{hamdi2020advpc}) or applying subtle geometric transformations to deceive models (TSI\cite{zhao2020isometry}). 
The intricate nature of these attack strategies, ranging from subtle geometric manipulations to query-based model probing, poses a significant challenge, demanding defense mechanisms that are not only robust but also generalizable across different attack modalities.

\subsection{Algorithm Details}

For clarity, we provide a detailed pseudocode of our Multimodal Robust Prompt Distillation (MRPD) framework in Algorithm~\ref{alg:mmpd}. The algorithm outlines the two key phases: the distillation phase, where multimodal knowledge is transferred into learnable prompts guided by a confidence-gated and dynamically weighted loss, and the lightweight inference phase, which utilizes the optimized prompts to achieve robust classification with zero additional computational cost.

\begin{algorithm}[h!] 
\newcommand{\KwParam}{\textbf{Parameters:}\hspace{0.5em}}
\newcommand{\KwHyper}{\textbf{Hyperparameters:}\hspace{0.5em}}
\KwIn{Training set $\mathcal{D} = \{(P, y)\}$, class names $T$}
\KwData{Student encoder $f_{\text{stu}}$, text encoder $f_T$, image teacher $f_I$, point teacher $f_P$, projection $\Pi(\cdot)$}
\KwParam{Point prompt $\theta_p$, text prompt $\theta_t$, dynamic weights $\lambda_I,\lambda_P,\lambda_T$}
\KwHyper{Temperature $\tau$, top-$k$ gate $k$}
\vspace{0.3em}
\hrule
\vspace{0.3em}
\textbf{Distillation phase}\;
\While{not converged}{
  Sample mini-batch $\{(P_i,y_i)\}_{i=1}^B \!\sim\! \mathcal{D}$\;
  $z_I \leftarrow \mathrm{normalize}\big(f_I(\Pi(P))\big)$ \tcp*{Image teacher}
  $z_P \leftarrow \mathrm{normalize}\big(f_P(P)\big)$ \tcp*{Point teacher}
  $z_T \leftarrow \mathrm{normalize}\big(f_T(\mathrm{concat}(V_{\mathrm{ctx}}(\theta_t), T))\big)$ \tcp*{Text teacher}
  $z_{\mathrm{stu}} \leftarrow \mathrm{normalize}\big(f_{\mathrm{stu}}(P; \theta_p)\big)$\;
  
  \tcp{Compute gated losses}
  $L_I \leftarrow \mathrm{ConfidenceGatedLoss}(z_{\mathrm{stu}}, z_I, z_T, y, \tau, k)$\;
  $L_P \leftarrow \mathrm{ConfidenceGatedLoss}(z_{\mathrm{stu}}, z_P, z_T, y, \tau, k)$\;
  $L_T \leftarrow \mathrm{ConfidenceGatedLoss}(z_{\mathrm{stu}}, z_T, z_T, y, \tau, k)$\;
  
  \tcp{Weighted sum with dynamic weights}
  $L_{\mathrm{total}} \leftarrow \sum_{m \in \{I,P,T\}} \left(\exp(-\lambda_m) L_m + \lambda_m\right)$\;
  
  Update $\theta_p, \theta_t, \{\lambda_m\}$ via $\nabla L_{\mathrm{total}}$\;
}
Obtain optimized prompts $\theta_p^*, \theta_t^*$\;

\vspace{0.3em}
\hrule
\vspace{0.3em}
\textbf{Inference phase}\;
$z_T^{\mathrm{robust}} \leftarrow \mathrm{normalize}\big(f_T(\mathrm{concat}(V_{\mathrm{ctx}}(\theta_t^*), T))\big)$\;
$z_P^{\mathrm{robust}} \leftarrow \mathrm{normalize}\big(f_{\mathrm{stu}}(P_{\mathrm{test}}; \theta_p^*)\big)$\;
$\hat{y} \leftarrow \arg\max_j \big( z_P^{\mathrm{robust}} \cdot z_T^{\mathrm{robust}}[j] \big)$\;
\Return $\hat{y}$\;
\caption{Multimodal Robust Prompt Distillation (MRPD)}
\label{alg:mmpd}
\end{algorithm}

\section{Extended Ablation Studies}

\subsection{Detailed Analysis on the Effectiveness of Multi-Modal Prompts}

\begin{table*}[h!]
\centering
\sisetup{detect-weight, mode=text, table-format=2.2, table-number-alignment=center}
\setlength{\tabcolsep}{4.5pt} 

\begin{tabular}{cc S S S S S S S S S S}
\toprule
\multicolumn{2}{c}{\textbf{Prompt Size}} & & \multicolumn{9}{c}{\textbf{White-Box Attack Performance}} \\
\cmidrule(lr){1-2} \cmidrule(lr){3-12}
{\textbf{Point}} & {\textbf{Text}} & {\textbf{Clean}} & {\textbf{PGD}} & {\textbf{Perturb}} & {\textbf{KNN}} & {\textbf{ADD-CD}} & {\textbf{ADD-HD}} & {\textbf{Drop-200}} & {\textbf{AdvPC}} & {\textbf{AOF}} & {\textbf{Avg. Robust}} \\
\midrule
5 & 3 & 89.71 & 87.93 & 76.66 & 86.35 & 78.97 & 64.26 & 78.97 & 36.47 & 49.64 & 69.91 \\
\rowcolor{gray!15}
\textbf{10} & \textbf{3} & \bfseries 90.52 & \bfseries 89.14 & \bfseries 80.79 & \bfseries 87.88 & \bfseries 81.36 & \bfseries 68.31 & \bfseries 78.32 & \bfseries 39.99 & \bfseries 54.86 & \bfseries 72.58 \\
10 & 5 & 89.71 & 88.70 & 78.85 & 87.24 & 79.38 & 65.24 & 77.55 & 38.37 & 50.93 & 70.78 \\
10 & 7 & 90.88 & 89.26 & 77.47 & 87.12 & 78.93 & 64.06 & 78.61 & 38.70 & 52.35 & 70.81 \\
15 & 3 & 89.38 & 88.41 & 78.81 & 86.67 & 79.70 & 63.78 & 76.74 & 35.70 & 49.72 & 69.94 \\
\bottomrule
\end{tabular}
\caption{Ablation study on the impact of prompt size on model performance under a comprehensive suite of white-box attacks. We evaluate different configurations of point and text prompt sizes on the ModelNet40 dataset. ``Avg. Robust'' is the average accuracy across all eight listed attacks. All accuracy values are in (\%). The best performing configuration is highlighted.}
\label{tab:ablation_prompt_size_full}
\end{table*}

To provide a comprehensive understanding of our framework's components, we present a detailed ablation study in Table~\ref{tab:ablation_white-box_full}, dissecting the individual and combined contributions of our proposed text and point prompts against the full suite of eight white-box attacks. The results for the Baseline (w/o Prompts) model first establish a critical performance datum, underscoring the profound vulnerability of standard VLM-guided models. On ModelNet40, its accuracy collapses to 0.00\% on four of the eight attacks, yielding a meager average robustness of 21.10\%. This degradation is even more pronounced on the real-world ScanObjectNN dataset, where the average robustness plummets to just 9.99\%, rendering the model entirely defenseless in adversarial conditions.
Introducing a single prompt type provides a substantial, albeit incomplete, defensive boost. The + Text Prompt variant excels at instilling semantic resilience, dramatically improving clean accuracy to 86.18\% on ModelNet40 and lifting average robustness to 52.37\%. This suggests that distilling knowledge into the text encoder anchors the model's predictions in a more stable semantic space. Conversely, the + Point Prompt proves slightly more effective at mitigating direct geometric perturbations on synthetic data, achieving a higher average robustness of 55.16\% on ModelNet40 with particular strength against Perturb and ADD-CD. An interesting dichotomy emerges on the more challenging ScanObjectNN dataset, where the text prompt is markedly superior to the point prompt (44.56\% vs. 35.67\% average robustness). This finding suggests that for noisy, occluded real-world data, reinforcing semantic stability is more critical than solely strengthening the geometric feature space.
The full MRPD model, which integrates both prompt types, demonstrates a powerful synergistic effect that far exceeds the additive contributions of its individual parts. On ModelNet40, the average robustness leaps to 72.58\%, an improvement of over 17 percentage points compared to the best single-prompt variant, achieving state-of-the-art performance across every attack category. This synergy is even more profound on ScanObjectNN, where the full model's average robustness of 67.39\% represents a fundamental shift in defensive capability, dramatically outperforming both single-prompt configurations. This detailed analysis unequivocally confirms that while each prompt offers partial protection, only their synergistic fusion, guided by our multi-modal distillation process, creates a truly comprehensive and robust defense resilient to a wide spectrum of adversarial manipulations.
\begin{table*}[h!]
\centering
\sisetup{
    detect-weight,      
    mode=text,          
    table-format=2.2,   
    table-number-alignment=center 
}
\setlength{\tabcolsep}{4.5pt} 

\begin{tabular}{
    l 
    S[table-format=2.2] 
    S[table-format=2.2] 
    S[table-format=2.2] 
    S[table-format=2.2] 
    S[table-format=2.2] 
    S[table-format=2.2] 
    S[table-format=2.2] 
    S[table-format=2.2] 
    S[table-format=2.2] 
    S[table-format=2.2, detect-weight] 
}
\toprule
 & \multicolumn{10}{c}{\textbf{White-Box Attack Performance}} \\
\cmidrule(lr){2-11}
{\textbf{Method}} & {\textbf{Clean}} & {\textbf{PGD}} & {\textbf{Perturb}} & {\textbf{KNN}} & {\textbf{ADD-CD}} & {\textbf{ADD-HD}} & {\textbf{Drop-200}} & {\textbf{AdvPC}} & {\textbf{AOF}} & {\textbf{Avg. Robust}} \\
\midrule
\rowcolor{gray!15}
{Full Model (Ours)} & \bfseries 90.52 & \bfseries 89.14 & \bfseries 80.79 & \bfseries 87.88 & \bfseries 81.36 & \bfseries 68.31 & \bfseries 78.32 & \bfseries 39.99 & \bfseries 54.86 & \bfseries 72.58 \\

{w/o Point Teacher} & 89.55 & 88.49 & 78.89 & 87.56 & 80.19 & 62.72 & 76.54 & 33.59 & 55.02 & 70.38 \\

{w/o Image Teacher} & 89.99 & 88.09 & 76.86 & 85.17 & 79.34 & 66.21 & 78.20 & 38.17 & 49.88 & 70.24 \\
\bottomrule
\end{tabular}
\caption{Ablation study on teacher models. This table compares the white-box attack performance of our full model against versions where either the point cloud teacher or the image teacher is removed. All results are for the \textbf{Multi-Modal Prompt} setting.}
\label{tab:ablation_teacher_white-box_full}
\end{table*}
We conduct a detailed ablation study to determine the optimal size for both the point and text prompts, with the results presented in Table~\ref{tab:ablation_prompt_size_full}. Our goal is to find a configuration that maximizes robustness without introducing excessive parameters that could lead to overfitting.
The results clearly indicate that a configuration with 10 point prompt tokens and 3 text prompt tokens achieves the best overall performance. This setup yields the highest average robustness of 72.58\% across all eight white-box attacks, while also maintaining a very high clean accuracy of 90.52\%.
When analyzing the impact of the point prompt size (while keeping the text prompt at 3 tokens), we observe that increasing the tokens from 5 to 10 provides a significant boost in average robustness (from 69.91\% to 72.58\%). This suggests that 10 tokens provide sufficient capacity to distill the complex geometric and structural knowledge from the teacher models. However, further increasing the point prompt size to 15 tokens results in a performance degradation across nearly all metrics, indicating that an overly large prompt may be harder to optimize or may begin to overfit.
Similarly, when we fix the point prompt at its optimal size of 10 tokens and vary the text prompt length, we find that a concise 3-token prompt is most effective. Increasing the text prompt to 5 or 7 tokens leads to a drop in average robustness. This suggests that a more compact semantic prompt creates a more stable and less noisy target space for distillation, facilitating a more effective alignment with the point cloud features. Based on this comprehensive analysis, we adopt the (10, 3) configuration for our final MRPD model.

\subsection{Ablation Study on Distillation Teachers}
To isolate and validate the contribution of each teacher modality, we conduct an ablation study where we systematically remove either the point cloud teacher or the image teacher from the distillation process. The results, presented in Table~\ref{tab:ablation_teacher_white-box_full}, demonstrate that both teachers provide unique and essential supervisory signals for achieving maximum robustness.
Our full MRPD model, which leverages the complete set of teachers (point, image, and text), achieves the highest average robustness of 72.58\%. When the Point Teacher is removed (w/o Point Teacher), the average robustness drops significantly to 70.38\%. This performance degradation is particularly pronounced on attacks that manipulate the fine-grained geometry of the object, such as ADD-HD (a drop of 5.59\%) and AdvPC (a drop of 6.40\%). This is expected, as the point teacher's role is to provide a high-quality geometric anchor based on the clean shape. Without this guidance, the student model is less able to maintain structural integrity against direct geometric manipulation.
Similarly, removing the Image Teacher (w/o Image Teacher) also leads to a substantial drop in average robustness to 70.24\%. The impact is most noticeable on attacks involving point-wise perturbations like Perturb (a drop of 3.93\%) and KNN (a drop of 2.71\%), as well as outlier-based attacks like AOF (a drop of 4.98\%). This highlights the image teacher's critical role in providing a holistically stable representation. Because 3D adversarial perturbations often lose their effectiveness when projected into 2D depth maps, this cross-modal teacher helps the model learn features that are invariant to such noisy manipulations.
In conclusion, this study confirms that the different teacher modalities are not redundant but rather complementary. The point teacher enforces geometric fidelity, while the image teacher provides global, perturbation-invariant supervision. The superior performance of the full model demonstrates that the synergistic fusion of these diverse and robust knowledge sources is essential for building a comprehensive defense.

\subsection{Qualitative Analysis on ScanObjectNN Dataset}
\begin{figure*}[t]
    \centering
    \includegraphics[width=\linewidth]{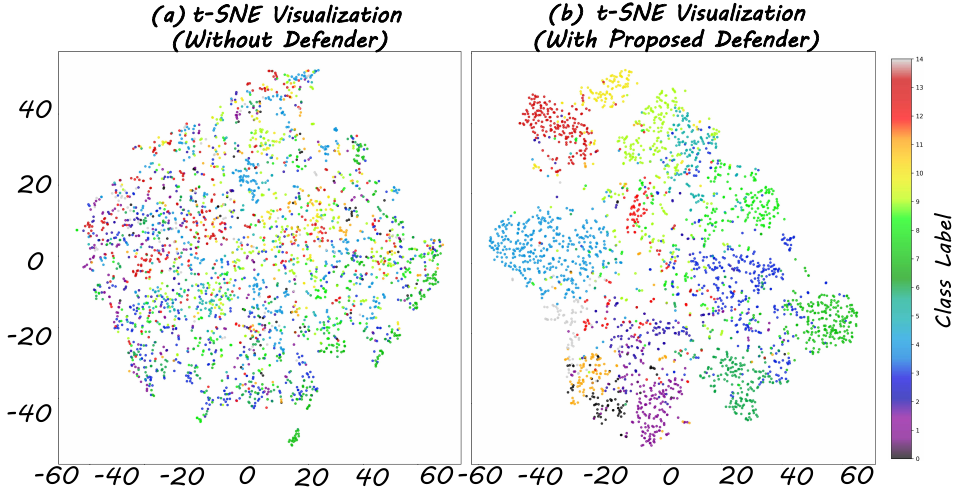} 
    \caption{
        t-SNE visualization of point cloud feature embeddings on the ScanObjectNN testset under the white-box Perturb attack. 
        \textbf{(a)} Features from the unprotected model collapse into an indistinguishable mass. 
        \textbf{(b)} Our method, MRPD, successfully restores clear class separation, leading to robust classification.
    }
    \label{fig:tsne_scanobjectnn}
\end{figure*}
To provide a qualitative understanding of our method's resilience, we visualize the feature embeddings on the challenging real-world \textbf{ScanObjectNN} dataset under the severe white-box \textbf{Perturb} attack.Figure~\ref{fig:tsne_scanobjectnn}(a) reveals the catastrophic impact of the attack on the unprotected model. The feature space has completely collapsed; embeddings from all classes are severely entangled, forming an undifferentiated mass at the center. This visual chaos directly explains the model's \textbf{0.00\%} accuracy under this attack, indicating a total loss of semantic structure.
In stark contrast, Figure~\ref{fig:tsne_scanobjectnn}(b) shows how our MRPD framework actively reorganizes this corrupted feature space. Distinct class manifolds emerge from the chaos, with features of the same category (e.g., light green, pink, teal) grouping into coherent and well-separated clusters. This re-established semantic structure is the direct reason for the model's dramatic performance recovery to \textbf{72.24\%} accuracy. This visualization provides compelling evidence that MRPD's robustness stems from its ability to preserve the fundamental integrity of the feature space, even when faced with strong attacks on noisy, real-world data.

\section{Experiments details}

\subsection{Implementation Details}
To ensure the reproducibility of our results and facilitate a fair comparison, our experimental protocol strictly follows the public benchmark established by IF-Defense, using their official implementation and default parameters for all baseline defenses and adversarial attacks. For data preprocessing, we uniformly sample 1,024 points for \textbf{ModelNet40} and 2,048 points for \textbf{ScanObjectNN} via Farthest Point Sampling, normalizing all point clouds into a unit sphere. Our MRPD framework is built upon a Point-CLIP V2 architecture, where only the prompts and dynamic loss weights are trained. Specifically, we use a text prompt of 3 tokens and a point prompt of 10 tokens. The model is trained for 100 epochs on a mixed-data diet of clean and PGD-attacked point clouds. We use the AdamW optimizer with a learning rate of \(1 \times 10^{-3}\) and weight decay of \(1 \times 10^{-2}\), coupled with a cosine annealing scheduler (T\_max=100, eta\_min=\(1 \times 10^{-5}\)). The learnable log-variance parameters for our dynamic loss were initialized to zero and clipped at a minimum value of -1.0 to ensure stability. All experiments were conducted on a workstation equipped with two NVIDIA RTX 4090 GPUs. The code for the baselines and attacks was sourced from the official IF-Defense repository (https://github.com/Wuziyi616/IF-Defense).

\subsection{Black-box Attack Data Generation}
Our black-box evaluation, presented in Table 2 of the main paper, employs a standard transfer-based attack methodology. This approach tests the generalizability of a defense by generating adversarial examples on a separate, known model (the ``surrogate model'') and then using these examples to attack the target models, which are treated as black boxes.
For our experiments, we selected PointNet++ as the surrogate model. PointNet++ is a widely recognized and powerful architecture for point cloud classification, and its distinct hierarchical feature learning mechanism makes it an excellent candidate for testing attack transferability to other architectures, including our own.
The procedure for generating the black-box attack dataset was as follows:
\begin{enumerate}
\item We first trained a standard PointNet++ model on the training sets of both ModelNet40 and ScanObjectNN until it achieved strong baseline performance.
\item Using this fully trained PointNet++ model, we then applied the complete suite of white-box attacks (PGD, Perturb, KNN, etc.) to generate adversarial point clouds for the entire test set of each dataset.
\item These resulting adversarial examples were saved and subsequently used as a static black-box test set to evaluate all models and defense methods reported in our study.
\end{enumerate}
\begin{table}[h!]
\centering
\setlength{\tabcolsep}{4pt} 
\begin{tabular}{l c c}
\toprule
\textbf{Attack Type} & \textbf{ModelNet40} & \textbf{ScanObjectNN} \\ 
\midrule
Clean       & 92.06 & 81.51 \\
\midrule
PGD         & 89.79 & 71.20 \\
Perturb     & 48.82 & 20.58 \\
KNN         & 0.08  & 0.00  \\
ADD-CD      & 52.27 & 46.74 \\
ADD-HD      & 34.89 & 29.81 \\
Drop-200    & 88.61 & 82.58 \\
AdvPC       & 3.77  & 4.41  \\
AOF         & 1.46  & 1.25  \\
\bottomrule
\end{tabular}
\caption{Performance of the PointNet++ surrogate model on the generated black-box attack dataset. Accuracy values are reported in percent (\%).}
\label{tab:surrogate_performance}
\end{table}
To validate the potency of the generated adversarial data, we measured the attack success against the surrogate PointNet++ model itself. As shown in Table~\ref{tab:surrogate_performance}, the attacks were highly effective, significantly degrading the model's accuracy. For several attack types, such as KNN and AOF, the accuracy dropped to nearly zero, confirming that the generated perturbations are strong and effective. This setup ensures that our black-box evaluation provides a rigorous and realistic test of our defense's ability to withstand attacks that were not directly optimized against its own architecture.
 
\end{document}